# Traffic Flow Combination Forecasting Method Based on Improved LSTM and ARIMA

## Boyi Liu

College of Information Science & Technology, Hainan University, Haikou, 570228

University of Chinese Academy of Science, Beijing, 100000

## Xiangyan Tang* and Jieren Cheng*

tangxy36@163.com*, cjr22@163.com*

*Corresponding author

College of Information Science & Technology, Hainan University, Haikou, 570228

State Key Laboratory of Marine Resource Utilization in South China Sea, Haikou, 570228

## Pengchao Shi

Mechanical and Electrical Engineering College, Hainan University, Haikou, 570228

**Abstract:** (ABS)

Traffic flow forecasting is hot spot research of intelligent traffic system construction. The existing traffic flow prediction methods have problems such as poor stability, high data requirements, or poor adaptability. In this paper, we define the traffic data time singularity ratio in the dropout module and propose a combination prediction method based on the improved long short-term memory neural network and time series autoregressive integrated moving average model (SDLSTM-ARIMA), which is derived from the Recurrent Neural Networks (RNN) model. It compares the traffic data time singularity with the probability value in the dropout module and combines them at unequal time intervals to achieve an accurate prediction of traffic flow data. Then, we design an adaptive traffic flow embedded system that can adapt to Java, Python and other languages and other interfaces. The experimental results demonstrate that the method based on the SDLSTM - ARIMA model has higher accuracy than the similar method using only autoregressive integrated moving average or autoregressive. Our embedded traffic prediction system integrating computer vision, machine learning and cloud has the advantages such as high accuracy, high reliability and low cost. Therefore, it has a wide application prospect.

traffic flow forecasting; LSTM; embedded system; depth learning

**Biographical notes:** (ABS) Mr. Boyi Liu, born in 1995. He is a B.S. student in the College of Information Science and Technology, Hainan University. His research interests involve machine learning, robotic, cloud compute.

Tang Xiangyan was born in 1981. She received the M.S. degree from School of Computer, Hunan Agricultural University in 2011. Now he is a lecturer at Hainan University. His research interests include artificial intelligence, network security, intelligent transportation, etc.

Dr. Jieren Cheng, born in 1974. He is a professor in the College of Information Science and Technology Hainan University. His research interests involve artificial intelligence, pattern recognition, algorithm design, and information security.

Mr. Pengchao Shi, born in 1995. He is a B.S. student in the College of Mechanical and Electrical Engineering Hainan University. His research interests involve machine learning, transportation.

## 1 Introduction

With the development of social economy and transportation, the severely frequent traffic problems and challenges in the traditional mode of traffic have aroused worldwide attention. In recent years, many countries have invested a large number of resources to carry out the development of management and control technology for road transportation system. Intelligent Transportation System (ITS) has been developed rapidly (Alam M et al., 2006; Moral-Muñoz J A et al., 2016), which can achieve quite a few exciting functions (Peter M and Vasudevan S K. 2017). Accurate traffic flow forecast is prerequisites and key steps to realize ITS, which are conducive to improving the efficiency of transport operations and the quality of folks' travel. ITS is also important to alleviate the road congestion, reduce carbon emissions, conserve the energy and so on. Relying on the current and historical traffic flow data, especially with the rapidly development of big data and machine learning in recent years. Excellent ITS forecasts the traffic flow reasonably and designs the best route for vehicles, realizing the traffic's balanced distribution in the road network and improving road utilization. It can achieve higher value of utility with the longer time of prediction. This paper made research on this problem and proposed a combination prediction method based on the improved long short-term memory neural network and time series autoregressive integrated moving average model (SDLSTM-ARIMA) with higher accuracy and longer prediction time. To address this issue, we designed an adaptive embedded system for the algorithm.

Traffic flow is an important measurement of the traffic state. It refers to the number of vehicles through a road section during a period (Bright P W et al., 2015). The excellent traffic flow prediction algorithm can predict the traffic flow for a certain period earlier and accurately. Traffic flow is affected by multifarious factors, while being subject to noise and non-linear interference, thus its rule is difficult to grasp, especially in the short-term traffic flow forecast (Habtemichael F G and Cetin M, 2016), which has been a challenging problem. In recent years, a large number of traffic flow prediction algorithms have been proposed [Lv Y et al., 2015; Hou Y et al., 2015], it can be broadly divided into two categories according to their forecasting basis: one prediction model is based on mathematical statistics and traditional mathematical such as calculus (Romero D et al., 2016); The other is developed by the means that mainly combined with modern science, technology and methods (Lopez-Garcia P, 2016).

The first class certainly includes a variety of traffic flow prediction algorithms. One of the representative results is Ahmed and Cook's time-series model (Fan Na, 2012) in the traffic flow prediction field for the first time in 1979, which includes the auto regressive model (AR) (Kumar S V and Vanajakshi L, 2015), the moving average model (MA) (Moorthy C K and Ratcliffe B G, 1988), the auto regressive moving average model (ARMA) (Williams B et al., 1998) and so on. The technology is mature and has high accuracy when the sample data is sufficient. It is used in relatively stable traffic, but subject to random factors. This method has high requirements for data and needs a large number of uninterrupted data. Stephanedes proposed history-average model (Stephanedes Y J et al., 1981) applied to urban traffic control system in 1981.This algorithm is simple and fast, but can not cope with emergencies; Okutani and Stephanedes proposed Kalman Filtering Model (Okutani I and Stephanedes Y J., 1984) for the traffic flow's prediction in 1984. Predictive factor selection in the algorithm is flexibility and this method has high precision and excellent robustness. But it requires a large number of matrix calculations and its forecast value is delayed for several time periods occasionally, which making it difficult to realize real-time online prediction. In addition, a series of traffic flow forecasting methods have been proposed in recent years, spatial-temporal characteristics-based analysis (Misra C et al., 2009), random forest model (Teles J, 2013) and similarity model (Hou Z and Li X, 2016), etc.

One of the representative traffic flow prediction algorithm of the second class is Davis and Nihan's Nonparametric Regressive Model (Davis G A and Nihan N L, 1991) applied to traffic flow prediction in 1991. Without prior knowledge, it can perform more accurate than parametric modeling only with sufficient historical data, but its complexity is also high. Dougherty proposed neural network (Dougherty M, 1995) for traffic flow prediction in 1995, which is suitable for complex and non-linear conditions. It can be effective to predict when the data is incomplete and inaccurate with outstanding adaptability and fault-tolerance. But it requires a great deal of learning data and the training process is complex. The classification regression tree method (Xu Yanyan et al., 2013) for the traffic flow forecast proposed by Xu Yanyan et al. in 2013. It has a better prediction effect and interpret ability but requires a great deal of training data and certain skills for parameter adjustment. In addition, plenty of traffic flow forecasting methods based on the above methods, deep belief network model (Liu F et al., 2015), support vector machine (Bai C et al., 2015), wavelet neural network model (Yanchong C et al., 2016), hybrid neural network model (Moretti F et al., 2015) have been proposed in recent years.

The traffic flow forecasting model proposed above has certain improvement in accuracy. Nevertheless, its prediction time of high precision is limited to 5min ~ 15 minutes while its prediction accuracy is not extremely high during 30min ~ 60min, and its stability is also poor. This work address this problem by developing an unequal interval combining model based on improved LSTM (Tian Y and Pan L, 2016), which can guarantee a high accuracy rate under the condition of reducing the prediction hours and the increasing the length of the sampling time interval.

The embedded system of traffic flow detection is the key to the realization of any traffic flow prediction algorithm. At present, there is still almost no complete and intelligent traffic flow detection system in the world. There is only a small number of discrete traffic flow detection systems (Lin H Y et al., 2017), and there is no connection between different detection points. To make the proposed traffic flow prediction algorithm based on machine learning come true. A distributed real-time updating and adaptive traffic flow monitoring system is designed in the work.

The rest of this paper is organized as follows. Section 2 proposes a traffic prediction algorithm based on long short-term memory (LSTM) neural network. Section 3 presents the design of the adaptive traffic flow prediction embedded system. Section 4 is the experiments of the traffic prediction algorithm. A spatial clustering process on the obtained lighting objects.

## 2 Improvement of LSTM neural network

### 2.1 LSTM Neural Network

The LSTM neural network is a special type of recurrent neural networks (RNN). RNN is an efficient and accurate depth neural network, which has outstanding effect in long-term dependence on data learning. RNN has been applied well in the field of machine translation, pattern recognition and so on. Nevertheless, it has a gradient disappearance problem. The LSTM neural network address this problem in RNN, which is characterized by the ability to learn long-term dependency information.

In the model, if the input sequence is set to $(x_1, x_2, \ldots, x_T)$ and the state of the hidden layer is set to $(h_1, h_2, \ldots, h_T)$, then at time $t$, there are:

$$i_t = sigmoid(W_{hi} h_{t-1} + W_{xi} X_t) \tag{1}$$

$$f_t = sigmoid(W_{hf} h_{t-1} + W_{hf} X_t) \tag{2}$$

$$o_t = sigmoid(W_{ho} h_{t-1} + W_{hx} X_t + W_{co} c_t) \tag{3}$$

$$c_t = f_t \odot c_{t-1} + i_t \odot \tanh(W_{hc} h_{t-1} + W_{xc} X_t) \tag{4}$$

$$h_t = o_t \odot \tanh(c_t) \tag{5}$$

In the formulas above, $i_t$ represents input gate, $f_t$ represents forget gate, $o_t$ represents output gate, $c_t$ represents cell, $W_h$ represents the weight of recursive link, $W_x$ represents the weight from the output layer to the hidden layer, the activation functions are sigmoid and *tanh*.

### 2.2 LSTM Neural Network Based on Self-Adaptive Probabilities

To address the over-fitting problem of natural network, Hinton proposed a solution named dropout in 2014 (Srivastava N, 2014). The "Dropout" method discards the neural network unit from the network temporarily according to a certain probability during the training process of the depth learning network. That is, dropout randomly selects a part of the neurons, then sets its output as 0, and remains its previous values at the same time. It restores the previous retention value in the next training process, and then randomly selects as well as repeats this process. In this way, the network structure changes in each training process. To avoid the situation that a feature is effective only with the support of the specific characteristics of other features, thus reducing the probability of over-fitting in the training process.

Although Hinton, et al. proposed dropout to reduce the probability of over-fitting. But they do not explore the calculation method further of the key parameter involved in dropout - the probability of selective discarding neurons, while they use the empirical value of 0.5. The reason is that the network structure generated randomly is the most in this case. In recent years, the empirical value is also used in the related applications based on LSTM. To address this problem, the work made a study and proposed the method of calculating the probability value of selective discarding neurons in dropout to improve the self-adaptive over-fitting of LSTM neural network.

In the improved scheme proposed in the work, the probability value of selective discarding neurons is replaced by the traffic data time singularity ratio. The reasons are as follows:

$$\frac{N_d}{N_j} = \frac{N_q}{N} \tag{6}$$

$$N_j = N_d + N_u \tag{7}$$

It can be deduced from (6), (7) that:

$$\frac{N_{qd}}{N_j} \leq \frac{N_d}{N_j} \tag{8}$$

In the formulas above, $N_d$ represents the number of discarded nodes, $N_j$ represents the number of nodes of each layer, $N_q$ represents the number of singular points, N represents the number of all nodes in the single-layer network, $N_u$ represents the number of nodes that is not discarded in the single-layer network, $N_{qd}$ represents the number of nodes in the single-layer network that are both discarded and belong to noise. The improved method proposed in the work can make the probability of selecting the node needed to delete randomly in dropout more reasonable. It is effective to prevent over-fitting problem is more prominent. We name the improved neural network SD-LSTM.

### 2.3 Traffic Flow Predicting Method Based on SDLSTM

This work applied SDLSTM into traffic flow prediction according to its features of great learning capacity on long term dependence of data and anti-overfitting.

As shown in Figure 1, the traffic flow prediction result is obtained by using the improved LSTM model. From the figure, we can see that most of MAPE values are under 10 in the time of heavy traffic flow. Nevertheless, the traffic flow base of 6 o'clock was high. Aiming at this, SDLSTM-ARIMA model is raised in the work to predict the traffic flow.

**Figure 1**  Traffic flow prediction error distribution based on improved LSTM

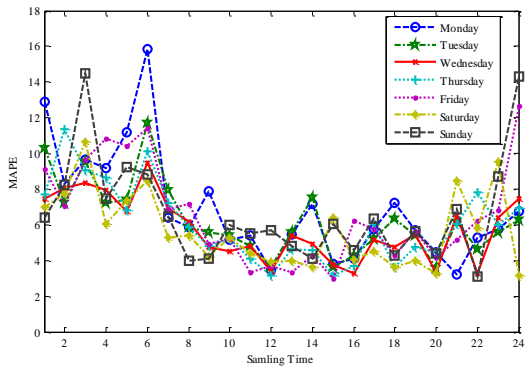

### 2.4 LSTM Neural Network Based on Self-Adaptive Probabilities

The work found that the main reason led to high the mean absolute percentage error (MAPE) in 6 o'clock is that the traffic flow changed severely during this period. The complicated features and high real time meant that LSTM cannot learn the whole features of this time slot. The work addresses the problem of the non-ideal result of 6 o'clock predication by bringing in traffic flow prediction method based on autoregressive integrated moving average (ARIMA). ARIMA doesn't have the training process of data learning, therefore, it is much more suitable for shorter period prediction. And the result is not ideal in the medium and long-term prediction. Aiming at this problem, the work solved it by combining LSTM and ARIMA with non-equal interval, that is the non-equal interval traffic flow prediction method based on SDLSTM neural network and ARIMA (a.k.a. SDLSTM-ARIMA).

Non-equal interval, that is, in the prediction period of LSTM, regarding 1 hour as unit time; in the prediction period of ARIMA, regarding 15 min as unit time. Under this circumstance, the traffic flow prediction in different periods during one day forms the condition of the combination of non-equal intervals.

The above is all of the methods. To realize it in reality, a comfortable embedded system is indispensable.

## 3 Design of adaptive traffic flow prediction embedded system

According to the traffic flow prediction algorithm, the design of adaptive traffic flow prediction embedded system is also carried out in this paper.

### 3.1 Working principle and system composition

The adaptive ability of the embedded system is realized by different traffic flow detection nodes. While we upload the traffic flow prediction model to the corresponding nodes of each embedded monitoring device in the cloud platform, we also upload a data training model. Each monitoring node will continue to upload data to the cloud platform while detecting traffic flow, and continuously train new data to improve the accuracy of the model. The traffic flow characteristics of different nodes are different, so the traffic flow prediction model is different. The embedded system designed by this way not only improves the accuracy of traffic flow prediction, but also realizes self-adaptation to different road traffic conditions. With the increase of the working hours of the embedded devices corresponding to each traffic monitoring node, the traffic data of the monitored road will be increasing, and the accuracy of the traffic flow prediction model trained by it will also increase.

For the hardware implementation of embedded devices, the development board used is raspberry PI 3, the camera used is the Raspberry Pi 3 Camera Module, the cloud platform used is the Ali cloud platform, and the traffic detection algorithm adopted is the vertical virtual detection line algorithm.

**Figure 2**  The composition of the embedded system

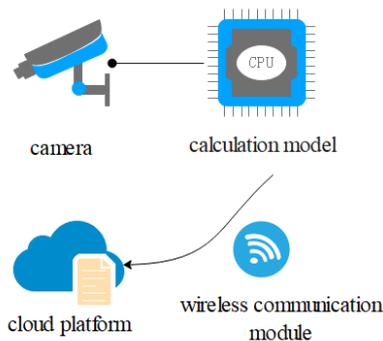

As shown in the Figure 2, It is the composition of the embedded system. Camera collects traffic video to the raspberry PI. The raspberry PI transfer data to the cloud platform by the wireless communication modules.

In the design of embedded systems, raspberry PI (RPi) is installed Linux system, python runtime environment already exists. RPi uses an ARM11 CPU running in 700MHZ. RPi can run a complete operating system, such as the common Linux distribution - RPi Distributions, such as Debian. This means that you can use your proficient language (such as Python, Java) and familiar libraries to develop, while running multiple processes in the background is also unstressed. RPi has a more comprehensive interface, and USB-host, RJ45, HDMI, SD card reader and other common interfaces are available. The data processing platform adopted is Storm. The past decade has been a decade of data processing revolutions. MapReduce, Hadoop, and related technologies have allowed us to process more data than ever before. Nevertheless, these data processing technologies are not real-time systems—they are not designed for real-time computing.

Therefore, large-scale, real-time data processing has increasingly become a business need. The lack of a "real-time version of Hadoop" has become a huge deficiency in the entire ecosystem of data processing. Storm fills this gap. Before Storm appears, you need to manually maintain a network of real-time processing the message queue (Queues) and message handler (Workers) thereof. The message handler takes a message from the message queue for processing, updates the database, and sends the message to other queues for further processing. Nevertheless, the limitations of this calculation method are too large, complex, not robust, and poor in scalability.

Twitter Storm is a distributed, real-time computing system for streaming computing. It is said to be real-time because of the way Storm handles data: Storm reads data streams directly from middleware and then goes directly to the memory for processing. Unlike Hadoop need to store it in the HDFS distributed file system, and then read into memory for processing punch the HDFS. This processing model of Hadoop is also referred to as "batch" mode. It is obviously suitable for processing large amounts of data that already exists, such as a year of weather data for analysis.

The adaptive traffic flow detection system designed in this paper is a system that needs real-time data processing. It needs to process traffic data in real time and update the database in real time (Srivastava N, 2018). At the same time, the amount of traffic data is very large. These features are all in accordance with the advantages of Storm, therefore, we chose Storm.

The traffic congestion detection algorithm used in the embedded system we designed here is the vertical virtual detection line algorithm (Jieren Cheng et al., 2018). The advantage of this algorithm compared to other algorithms is that it does not require installation of geomagnetic coils and other testing equipment. It can be achieved only by using video processing. In addition, compared with the horizontal virtual detection line algorithm, it can effectively avoid the omission of the vehicle, making the detection result more accurate. In the vertical virtual detection line algorithm, it sets a vertical virtual detection line for traffic flow detection in the middle of the road. Image segmentation based on the hybrid Gaussian model is applied to every frame of video, and then the pixel gray information about vertical detection line in every frame is extracted continuously. The pixel gray distribution of the algorithm detection line is converted into a histogram. When the program loads every frame of the video continuously, the histogram changes continuously. The algorithm calculates the number of sudden changes in the unit time of a histogram as the number of vehicles passing in a unit time.

### 3.2  Work flow of the system

Before installing the equipment, we need to upload the traffic forecast model and the data training model to the cloud platform. During the work of the equipment, the camera collects traffic video of the road. Then it uses the vertical virtual detection line algorithm to detect traffic flow per unit time. After the equipment collects the video data, the data will be uploaded to the cloud platform that has a traffic flow prediction model. The platform carries out the traffic forecast program, we proposed in the paper, predicts the traffic flow data at the next unit time, and returns to the user. At the same time, the new obtained data will be used to train the prediction model and improve the accuracy.

**Figure 3**  Data flow diagram of the system

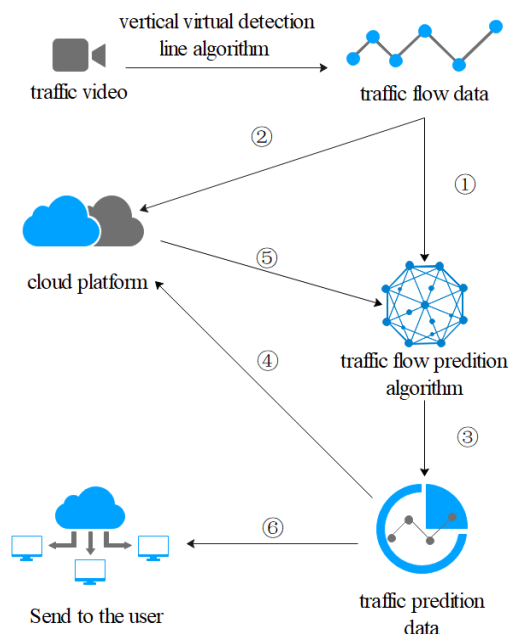

As shown in Figure 3, it is the data flow in the embedded system. Input the traffic video that collected from the camera and output the predation data.

The arrow 1 shows the data transmission process. The data is uploaded to the prediction model in the computing node directly. The mode of transmission is transmitted by wireless network. This is because traffic flow video has been transformed into integer data in embedded systems so the requirement for transmission is low. Each discrete embedded traffic flow detection node is equipped with a wireless communication module. The platform assigns ID numbers of different monitoring points. The data is uploaded to the platform from the monitoring point, and the platform assigns different modules to this data based on its ID. Therefore, the traffic flow data from different detection points can be uploaded to different computing modules in the platform.

The arrow 2 represents a storage module that uploads data to the cloud platform. It can achieve by the wireless model and some new protocols (Tang H et al., 2016). The platform needs to store all traffic data, including previous and the latest ones. For different traffic flow detection nodes, different calculation modules are allocated. Because the traffic data of different nodes are different. This allocation method ensures the adaptive capability at the computing service level of the traffic flow prediction system. The cloud platform is mainly used for data storage, updating, and computing. It needs to receive the traffic flow data from the detection node in real time and update the traffic flow database of different nodes in real time. At the same time, it is indispensable to allocate the computing resources to the operation of the traffic flow prediction algorithm for different detection nodes. A large part of the computing resources of the platform are used for the training of the prediction models corresponding to each node. It is indispensable to give compute resource supports because each node corresponds to a different traffic flow prediction model.

The arrow 3 shows the prediction model calculates traffic flow data that needs to be predicted. The predation model is the algorithm that is shown in the work before. In the following experiments, it proved to be effective.

The arrow 4 indicates that the predicted data also need to be uploaded to the cloud platform. That is because the calculation model's update not only need the real traffic flow data of this detection monitor, but also need the prediction data. Calculation model updates itself by the real data and the prediction data. It realizes the adaptive ability in the calculated level.

The arrow 5 shows the processing that the cloud platform sends data and assigns computing resources to the calculation models. In this step, we can see the advantage of Storm, it can assign the resource to a large number of different compute models. By using Storm, it ensures that there are enough computing resources for different detection points. The calculation models of different detection points can also update themselves in real time with the support of the Storm.

The arrow 6 means the prediction data spread to all the users of the cloud platform. All the users who have jurisdiction can use the data. It is beneficial to extensibility develop. As an example, the prediction of traffic flow based on the relation between different points at the same time, the traffic flow prediction of the whole city and so on. A large and complete traffic flow data are preserved in an orderly manner, which is of great value.

Compared with other traffic flow prediction embedded systems, the system designed in this paper combines computer vision, machine learning and cloud platform. This embedded system makes the prediction result more accurate. The robustness of the prediction model is gradually improved and it has the ability to adapt the video from different nodes.

Finally, experiments have been conducted to valid correctness and the practicability of our scheduling algorithm.

## 4 Experiment

### 4.1 Data and environment of the experiment

The experimental traffic data in the work come from the traffic flow data set published publicly on the official website of British Columbia of Canada (The Ministry of transportation and transportation in British Columbia, 2018).

The experiment of the work is implemented by python 3.5.2 language in the 2-core 4-thread, 2.5GHz, 8G memory computer with Linux system (Ubuntu 16.04 64-bit), where the traffic flow is collected every 1 hour.

### 4.2 Result and Analysis of SDLSTM-ARIMA Traffic Flow Prediction

The formula below shows the method of calculating the MAPE value of the data deviation. As shown in Formula 9:

$$MAPE = (\sum \frac{\frac{X-Y}{X}}{N}) \times 100\% \div N \qquad (9)$$

The improved LSTM neural network for data training is used in the experiment, and the MAPE changes in process are shown in Figure 4. The ordinary LSTM neural network method obtained by training is compared with the SDLSTM-ARIMA method proposed in the work, as shown in Figure 5, where the red part of the figure shows the improvement of the accuracy of the prediction after the introduction of the ARIMA model. The result of the comparison proved the effectiveness of SDLSTM-ARIMA.

Finally, the SDLSTM-ARIMA model obtained from the training of the data has been tested in the test data in the experiment. Figure 6 shows the forecast value and the actual value of one day of the results.

**Figure 4**  The change of MAPE with epoch during training

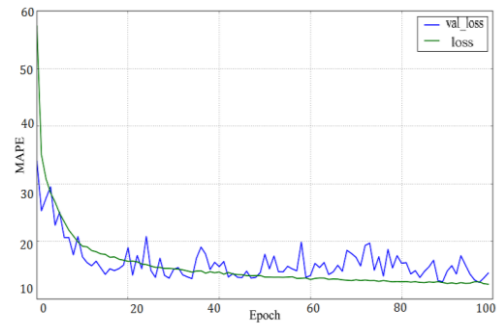

**Figure 5**  Comparison of predicted scenarios between SDLSTM-AR and LSTM

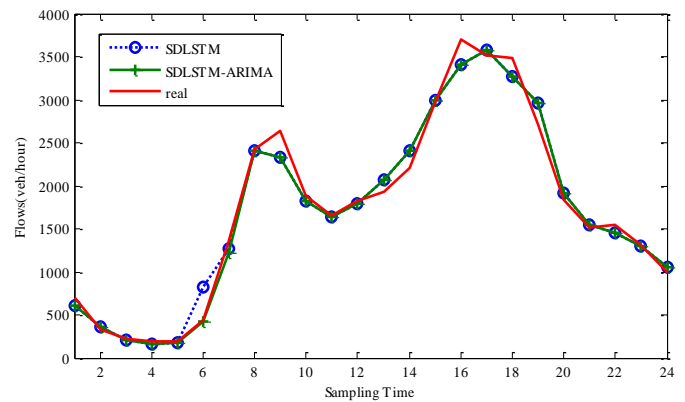

**Figure 6** The predicted values obtained by SDLSTM-AR method compared with the real values in a day

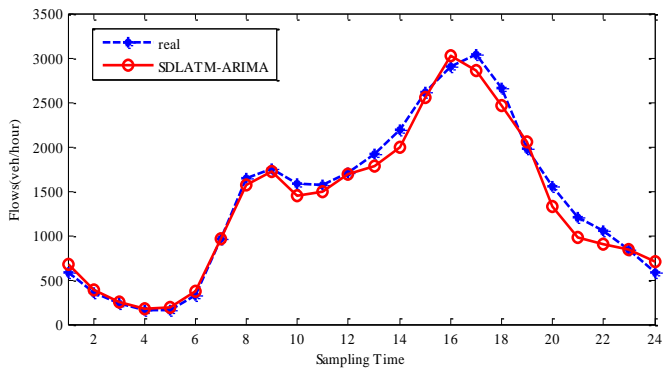

The work compared the SDLSTM-ARIMA method proposed in the work with the commonly used ARIMA prediction method and the latest proposed AR-RBLTFa method (Hou Z and Li X, 2016) by the experiments in the testing data set and training data set. Figure 7 shows the comparison of prediction results between different methods in working days. Figure 8 shows the comparison of prediction results between different methods in non-working days. As can be seen from the figures, compared with the commonly used ARIMA prediction method, AR-RBLTFa method and SDLSTM-ARIMA have higher accuracy.

**Figure 7** Comparison of experimental results of different methods of working days

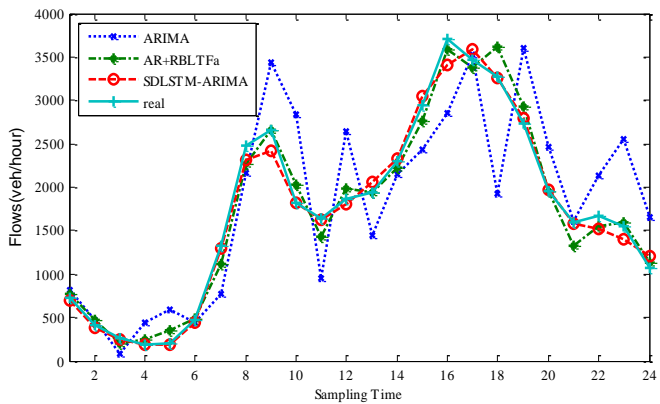

**Figure 8** Comparison of experimental results of different methods of non-working days

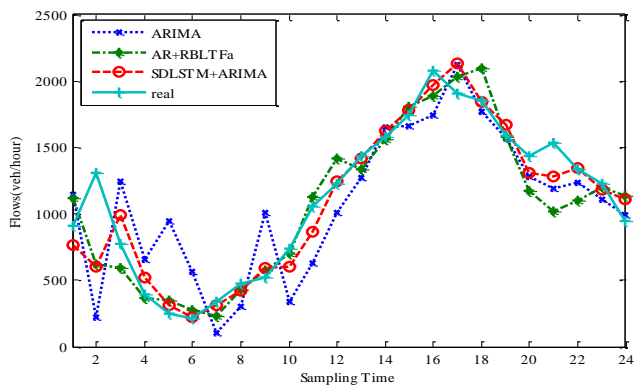

The accuracy rate in the experiment is: SDLSTM-ARIMA > AR-RBLTFa > ARIMA (as illustrated in Figure 8). Experiments show that the SDLSTM is effective and accurate.

## 5 Conclusion

The work addresses the problem that the traffic flow prediction algorithm cannot reach ideal result in the medium and long-time slot by developing the LSTM neural network. SDLSTM method is put forward. This work defined the calculation method of time singularity ratio of the traffic flow firstly, improved LSTM neural network and put forward the probability values of selectively discarding neurons of the dropout model by using time singularity ratio as self-adaptive data environment to deal with the problem of over-fitting in LSTM neural network and achieve adaptivity of the traffic flow data set. Then, this work applied SDLSTM neural network in the traffic flow prediction. At last, we verified the method by experiments and compared it with other methods. The result shows that SDLSTM-ARIMA proposed in this paper has higher accuracy and stability. This method converts the traffic big data to practical value by using big data technology and machine learning. And it has broad application prospects. We also designed an embedded system. The system combined with raspberry PI, cloud platforms, camera. We use Storm to achieve real time and distributed computation. This system can adopt different monitoring points and update the calculation model in real time. Combined with the traffic flow prediction algorithm and the self-adapting embedded system, we can get more accurate traffic flow prediction.

**Acknowledgments**

This work was supported by the National Natural Science Foundation of China (Project No. 61762033, 61363071, 61471169); the National Natural Science Foundation of Hainan (Project No. 617048, 2018CXTD333); Hunan Province Education Science Planning Funds (Project No. XJK011BXJ004); Hainan University Doctor Start Fund Project (Project No. kyqd1328); Hainan University Youth Fund Project (Project No. qnjj1444); State Key Laboratory of Marine Resource Utilization in the South China Sea, Hainan University; College of Information Science & Technology, Hainan University; Nanjing University of Information Science & Technology (NUIST); National Training Programs of Innovation and Entrepreneurship for Undergraduates (Project No. 201610589002). A Project Funded by the Priority Academic Program Development of Jiangsu Higher Education Institutions; and the Jiangsu Collaborative Innovation Center on Atmospheric Environment and Equipment Technology.